\documentclass[10pt,letterpaper]{IEEEtran}
\usepackage{color}
\usepackage{url}
\usepackage[latin9]{inputenc}
\usepackage{epstopdf}
\usepackage{amsthm}
\usepackage{amsmath}
\usepackage{amsthm}
\usepackage{amssymb}
\usepackage{graphicx}
\usepackage{wrapfig}
\usepackage{mathrsfs}
\usepackage{float}
\usepackage{mathtools}
\usepackage{tabulary}
\usepackage{booktabs}
\usepackage{caption}
\graphicspath{{./}{./files/figures_eps/}}
 
\usepackage{subfig} 
\usepackage{booktabs} 
\usepackage[linesnumbered,ruled,vlined]{algorithm2e}

\usepackage{makecell} 

\PassOptionsToPackage{bookmarks={false}}{hyperref}

\IEEEoverridecommandlockouts

\def\BibTeX{{\rm B\kern-.05em{\sc i\kern-.025em b}\kern-.08em
   T\kern-.1667em\lower.7ex\hbox{E}\kern-.125emX}}

\newcommand{\comment}[1]{ }

\newcommand{\dsone}{{\tt LoRa/In}} 
\newcommand{\dstwo}{{\tt LoRa/Out}} 
\newcommand{\stthree}{{\tt Wired}} 
\newcommand{\stfour}{{\tt WiFi}} 
\newcommand{\dsosu}{{\tt OSU-LoRa}} 
\newcommand{\dsne}{{\tt NE-Wired/WiFi}} 

\newcommand\subparagraph{%
  \@startsection{subparagraph}{0}
  {\parindent}
  {0ex \@plus 0ex \@minus 0ex}
  {-1em}
  {\normalfont\normalsize\bfseries}}
\makeatother

\usepackage{titlesec}
\titlespacing*{\section}
{0pt}{1.2ex}{0ex} 
\titlespacing*{\subsection}
{0pt}{1ex}{0ex}  
\titlespacing*{\subsubsection}
{0pt}{0ex}{0ex} 

\begin{document}
\bstctlcite{IEEEexample:BSTcontrol}


\title{An Analysis of Complex-Valued CNNs for RF Data-Driven Wireless Device Classification}


\author{Jun Chen${^\dag}$, Weng-Keen Wong${^\dag}$, Bechir Hamdaoui${^\dag}$, Abdurrahman Elmaghbub${^\dag}$, Kathiravetpillai Sivanesan${^\ddag}$, Richard Dorrance${^\ddag}$, Lily L. Yang${^\ddag}$~\\
 $^\dag$ \small Oregon State University, Corvallis, OR, USA; Email: \{chenju3,wongwe,hamdaoui,elmaghba\}@oregonstate.edu \\
  ${^\ddag}$ \small Intel Corporation, Hillsboro, OR, USA; Email: \{kathiravetpillai.sivanesan,richard.dorrance,lily.l.yang\}@intel.com \\
}
 
\maketitle
\thispagestyle{plain}
\pagestyle{plain}

\begin{abstract}
Recent deep neural network-based device classification studies show that
complex-valued neural networks (CVNNs) yield higher classification accuracy than
real-valued neural networks (RVNNs). Although this improvement is (intuitively)
attributed to the complex nature of the input RF data (i.e., IQ symbols), no
prior work has taken a closer look into analyzing such a trend in the context of
wireless device identification. Our study provides a 
deeper understanding of this trend using real LoRa and WiFi RF datasets. We perform a deep dive into understanding the impact of (i) the input representation/type and (ii) the architectural layer of the neural network. For the input representation, we considered the IQ as well as the polar coordinates both
partially and fully. For the architectural layer, we considered a series of
ablation experiments that eliminate parts of the CVNN components. Our results
show that CVNNs consistently outperform RVNNs counterpart in the various scenarios
mentioned above, indicating that CVNNs are able to make better use of the
joint information provided via the in-phase (I) and quadrature (Q) components of the signal.
\footnote{This work is supported in part by NSF/Intel Award No. 2003273.}
\footnote{\textregistered 2022 IEEE. Personal use of this material is permitted. Permission from IEEE must be obtained for all other uses, in any current or future media, including reprinting/republishing this material for advertising or promotional purposes, creating new collective works, for resale or redistribution to servers or lists, or reuse of any copyrighted component of this work in other works.}


\end{abstract}


\section{Introduction}
%
Radio frequency (RF) fingerprinting has emerged as a key enabler for providing automated device identification proven useful in a wide range of new wireless applications~\cite{Soltanieh2020ARO,Riyaz2018DeepLC,Restuccia2019DeepRadioIDRC}.
In essence, RF fingerprinting extracts features from transmitted RF
signals to distinguish between different devices~\cite{hamdaoui2020deep,elmaghbub2020widescan}.
Traditionally, RF fingerprinting relies on carefully
hand-crafted features that require domain knowledge of the underlying
communication (e.g., modulation) protocols. Recently, a
growing number of deep learning-based algorithms have been proposed for RF
wireless signal classification (see Section \ref{sec:related}). Unlike the
traditional approach, recent deep learning approaches learn a representation directly from the raw IQ samples of wireless signals and thus do not rely on any domain knowledge to construct features. These deep learning methods show strong benefits in applying neural networks to wireless signal classification tasks.

While much of the existing literature in this area has focused on using
real-valued neural networks (RVNNs)
\cite{Riyaz2018DeepLC,OShea2016ConvolutionalRM}, some researchers have shown
that complex-valued neural networks (CVNNs)
\cite{Gopalakrishnan2019RobustWF,Agadakos2019DeepCN}, which account for the
complex-valued aspects of the data, can be more effective. Although these
approaches show that CVNNs can outperform RVNNs, a deeper analysis providing
insights into why CVNNs are more effective than RVNNs is still lacking.

In this work, we study RVNNs and CVNNs vis-a-vis of their ability to identify
and classify wireless devices using RF signals collected from real-world
datasets capturing two widely used protocols, LoRa and WiFi. Specifically, we
compare the performances of CVNNs and RVNNs and perform a deeper analysis of the
benefits of CVNNs through a series of experiments, including ablation
experiments that remove parts of the CVNN architecture to understand which parts
of the network have the most impact on its predictive performance.
Our experimental results show that:
\begin{itemize}
\item CVNNs consistently outperform their "equivalent" (in terms of the number
  of neural network parameters) RVNNs on RF/device classification tasks under
  various experimental settings.
\item CVNNs are able to exploit the joint In-phase (I) and quadrature (Q)
  feature information of the RF signals more effectively than RVNNs.
\item In general, ablated CVNNs yield lower accuracy, which indicates that
  removing cross-terms causes information loss and makes CVNNs behave like
  RVNNs.
\item Ablated CVNNs exhibit decreasing trends in performance with higher layers
  being ablated, indicating that similar to RVNNs, deeper layers of CVNNs
  contain higher-level features that result in higher accuracy.
\item In some cases, removing part of the layer improves
  accuracy. This break in the decreasing trend makes us speculate that the
  additional cross-terms of CVNNs can not only enrich information but also
  increases the redundancy of information and harm the performance.
\end{itemize}

The remainder of this paper is organized as follows. Section \ref{sec:related}
describes the related works applying deep learning approaches on RF
signal/device classification. Section \ref{sec:data} describes the datasets used in our evaluation. Section \ref{sec:method} presents
the detailed neural networks configuration and ablation strategies used in this work. Experimental results are discussed in Section \ref{sec:expr} and Section \ref{sec:conc} concludes the paper.

\section{Related Work}
\label{sec:related}

Early works on RF fingerprinting focused on automatic modulation classification and on using model based approaches to hand-craft and design features~\cite{OShea2016ConvolutionalRM,West2017DeepAF,OShea2017AnIT}. 
More recent RF fingerprinting works have shifted towards using deep learning to extract features from RF signals automatically~\cite{Riyaz2018DeepLC,hamdaoui2020deep,bao2020iot,Sankhe2019ORACLEOR,Soltani2020MoreIB,basha2021leveraging}.
%
For instance, Soltani et al. \cite{Soltani2020MoreIB} proposed a
data augmentation method for raw IQ data samples that works without
needing prior knowledge about the waveform and the receiver-transmitter coordination.
The approach proposed in~\cite{hamdaoui2020deep,elmaghbub2020widescan} leverages out-of-band spectrum emissions that are caused by hardware impairments to enhance fingerprinting accuracy.
Al-Shawabka et al. \cite{AlShawabka2020ExposingTF} investigate the impact of the
wireless channel on fingerprinting accuracy.

Some approaches used CVNNs instead of RVNNs to enable fingerprinting~\cite{Gopalakrishnan2019RobustWF,Agadakos2020ChameleonsOC}. For example, Gopalakrishnan et al. \cite{Gopalakrishnan2019RobustWF} employed CVNN for
RF fingerprinting and showed the benefits of the preamble and
noise augmentation on the accuracy. 
The CVNN architecture used in \cite{Gopalakrishnan2019RobustWF} is a
hybrid neural network that inserts real-valued layers after complex
convolutional layers. Agadakos et al. \cite{Agadakos2020ChameleonsOC} developed
two novel deep CVNNs (DCN) that are agnostic of the underlying protocols.
They found that these CVNNs outperform their real-valued counterparts, even with
fewer parameters. However, the authors did not provide the details of their neural network models, nor did they provide a deeper analysis of why the CVNN is better
than RVNN. Unlike past work, our paper, for the first time, will
focus on a detailed analysis of why the CVNNs are a better choice for RF fingerprinting.

\section{RF Datasets}
\label{sec:data}
We next provide brief descriptions of the LoRa and WiFi RF datasets used for our evaluation, which are collected respectively at Oregon State University~\cite{elmaghbub2021lora} and at Northeastern University~\cite{AlShawabka2020ExposingTF}, with further details found in the corresponding references.


\subsection{LoRa Dataset Scenarios}
We consider two LoRa dataset scenarios, \dsone~and
\dstwo, collected using a testbed comprised of $25$ Pycom transmitters and USRP B$210$
receivers. For a detailed description of the full datasets, refer to~\cite{elmaghbub2021lora}.
\begin{itemize}
\item{\dsone:} This is an indoor environment scenario, where
  RF samples are captured from the $25$ Pycom devices, each transmitting the same message from the same location, $5$m away from the receiver. For more details, refer to Setup 1 in \cite{elmaghbub2021lora}.

\item{\dstwo:} To allow for performance evaluation while considering the
  impact of outdoor wireless channel impairments, we also considered an outdoor
 environment scenario. In this scenario too, all devices transmit the same message from the same location, $5$m away  from the receiver. For more details, refer to Setup 2 in \cite{elmaghbub2021lora}.
\end{itemize}

\subsection{WiFi Dataset Scenarios}
The previous two dataset scenarios focus on LoRa device fingerprinting. For
completeness,
we also consider the following two dataset scenarios: \stthree~and
\stfour~\cite{AlShawabka2020ExposingTF}.
\begin{itemize}
\item {\bf \stthree:} This dataset scenario (corresponding to Setup C
  in~\cite{AlShawabka2020ExposingTF}) is collected by Northeastern University
  (NE) team using a testbed of 20 SDR (software define radios), composed of 13
  N210 and 7 X310 USRPs. The data acquisition is done using a USRP N210 receiver
  with a sampling rate of
  $20$MS/s. 
  All transmitters were connected to the receiver using the same coaxial RF SMA
  cable and a 5db attenuator one at a time. By this setup all transmitters are
  not affected by multi-path and experience exactly the same channel conditions.

\item {\bf \stfour:} This dataset scenario (corresponding to Setup D
  in~\cite{AlShawabka2020ExposingTF}), also provided by the same NE team, is
  collected using a testbed is constituted of 10 SDRs (4 N210 and 6 X310 USRPs).
  The devices are located in an anechoic chamber and are connected to the same
  transmitting antenna. The data acquisition is done using a USRP N210 receiver
  with a sampling rate of $20$MS/s. Each device transmits the same IEEE $802.11$
  a/g packets for $30$s, one at a time.
\end{itemize}

\section{Methodology}
\label{sec:method}

We now describe our experimental methodology, which includes details on the
datasets used, the neural network architectures and the experiments conducted.

\subsection{Dataset Construction for Experimental Evaluation} \label{sec:dc}
Typically, an evaluation methodology like averaging over multiple random
train/test splits or cross-validation is used. However, RF data experiences
changes in the data distribution over the duration of the transmission. This
non-stationary behavior violates the assumptions of standard machine learning
evaluation methodologies. Data from different transmissions from the same device
can also exhibit quite a large variance. As a result, we need to create a
specialized methodology suitable for our RF data-based experiments.

\begin{figure}
  \centering \includegraphics[width=0.45\textwidth]{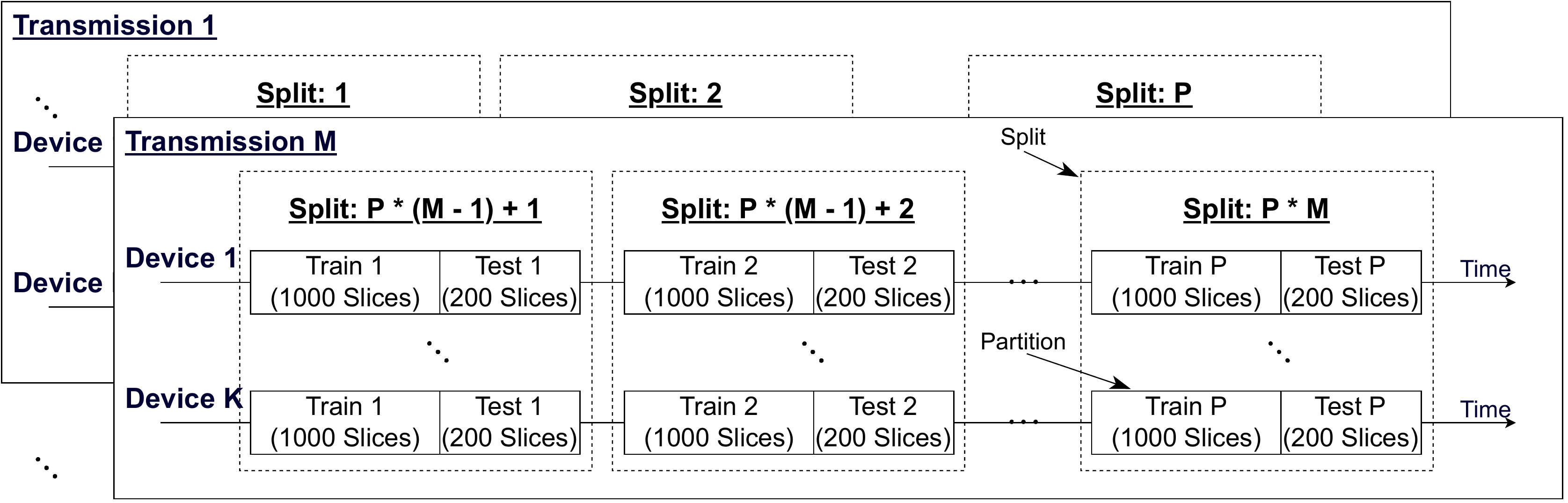}
  \caption{Constructing splits from the data} 
  \label{fig:ds_split}
\end{figure}

To mitigate the non-stationary behavior, we ensure that the training data is
close in time to the test data. In addition, for each device, we restrict the
training and test data to be from the same transmission. Consequently, we
create $S$ datasets which we refer to as \emph{splits} and evaluate the average
accuracy of these $S$ splits. Suppose there are $K$ devices (the number of
classes) and we only use $M$ transmissions per device in our experiments. To
form splits, we create $P$ train/test partitions in each transmission and these
partitions are created from evenly-spaced time intervals over the transmission
data. A split consists of the corresponding partitions of corresponding
transmissions over all devices, where \emph{corresponding} means the same index, as
shown in Figure \ref{fig:ds_split}. To be specific, the \emph{split} with index of
$P\times (m - 1) + p$ is the union of the partition $p$ in transmission $m$ of
all $K$ devices, where $m \in [1, ..., M]$ and $p \in [1, ..., P]$. Since we can
form $P$ splits per transmission over all devices, the total number of splits is
$S = P \times M$.


We create the splits using the middle third of the data to avoid dealing with
specialized data in the first or last third of the transmission. For one split,
there are $K$ partitions ($1$ partition from each of the $K$ devices). Each partition
contains $1200$ consecutive IQ samples at the beginning. We create $1200$
overlapping sliding windows based on these consecutive IQ samples of partition,
where we refer to the data within a window as a \emph{slice}. Thus, using the
first IQ sample of the partition as the starting point, we take a window of
$100$ samples as the first slice for a given device. Then, we advance the
sliding window on the transmission in which the partition is by a stride of size
$1$ and take the next $100$ samples as the second slice. This process continues
until the last sample of the partition (as the starting point of the last
slice). We repeat this process for each device. Each split is formed by taking
the union of the generated slices from partition $p$ from transmission $m$ over
all devices. The total number of slices from a split is $1200\times K$ and
each slice has a dimension of $2\times 100$ because the IQ sample has the two parts
of I and Q.

The ratio of training to test instances is $5:1$ and we ensure that the training
data has an equal number of slices from each device. The final results of our
experiments evaluate and show the accuracy averaged over all $S$ splits in each
datasets scenario. For \dsosu~dataset scenarios $K=25, M=1, P=50, S=50$ and for
\dsne~dataset scenarios $K=20, M=3, P=50, S=150$.


\subsection{Complex-Valued Neural Networks: CVNNs}
Early work on CVNNs used pure complex number operations and focused on the
topics of complex activation functions
\cite{Nitta1997AnEO,Georgiou1992ComplexDB} and complex backpropagation
\cite{Leung1991TheCB, Benvenuto1992OnTC}; this work was primarily intended for
shallow neural networks. Modern software packages for deep networks are largely
built around automatic differentiation for real values. Software to extend
automatic differentiation to complex values is not currently mature and often
suffers from numerical instability. As a result, modern software packages
attempt to reuse real-valued functionality to approximate complex-valued
building blocks such as complex-valued convolutional layer, complex-valued
ReLUs, and complex-valued batch normalization layer \cite{Trabelsi2018DeepCN}.

The most important component of a CVNN is the complex-valued convolutional layer
which reflects the essential difference between the CVNNs and RVNNs. Each
complex-valued convolutional block within the layer contains a complex filter matrix $W = A + iB$ and an
input vector $h = x + iy$; here, $A$ and $B$ are real matrices while $x$ and $y$
are real vectors. The complex convolution can be expanded according the
distributive property of convolution operator:
\begin{equation}
  \label{eq:ccov1}
  \mathbf{W} * \mathbf{h}=(\mathbf{A} * \mathbf{x}-\mathbf{B} *
  \mathbf{y})+i(\mathbf{B} * \mathbf{x}+\mathbf{A} * \mathbf{y})
\end{equation}
We can rewrite the above equation to matrix form as:
\begin{equation}
  \label{eq:ccov2}
  \left[\begin{array}{c}{\operatorname{Re}(\mathbf{W} * \mathbf{h})} \\ {\operatorname{Im}(\mathbf{W} * \mathbf{h})}\end{array}\right]
  = \left[\begin{array}{rr}{\mathbf{A}} & {-\mathbf{B}} \\ {\mathbf{B}} & {\mathbf{A}}\end{array}\right]
  * \left[\begin{array}{l}{\mathbf{x}} \\ {\mathbf{y}}\end{array}\right]
\end{equation}
Equation \eqref{eq:ccov1} shows that the complex-valued convolutional layer
needs two real-valued convolutional filters $A$ and $B$. For the complex-valued
vector $h$ with real part $x$ and imaginary part $y$, both the real and
imaginary part of the output of complex-valued convolution operator consist of
these four components ($A, B, x, y$). The only difference between the two is that the real part of
the output is the subtraction between $\mathbf{A} * \mathbf{x}$ and $\mathbf{B}
* \mathbf{y}$, and the imaginary part is the addition between the cross terms
$\mathbf{B} * \mathbf{x}$ and $\mathbf{A} * \mathbf{y}$. This interleaving
between the real and imaginary components is the most significant difference
between CVNNs and RVNNs.
  

\subsection{Neural Network Architectures} \label{sec:expr_nets}


\begin{figure}
  \centering \subfloat{%
    \includegraphics[width=0.45\textwidth]{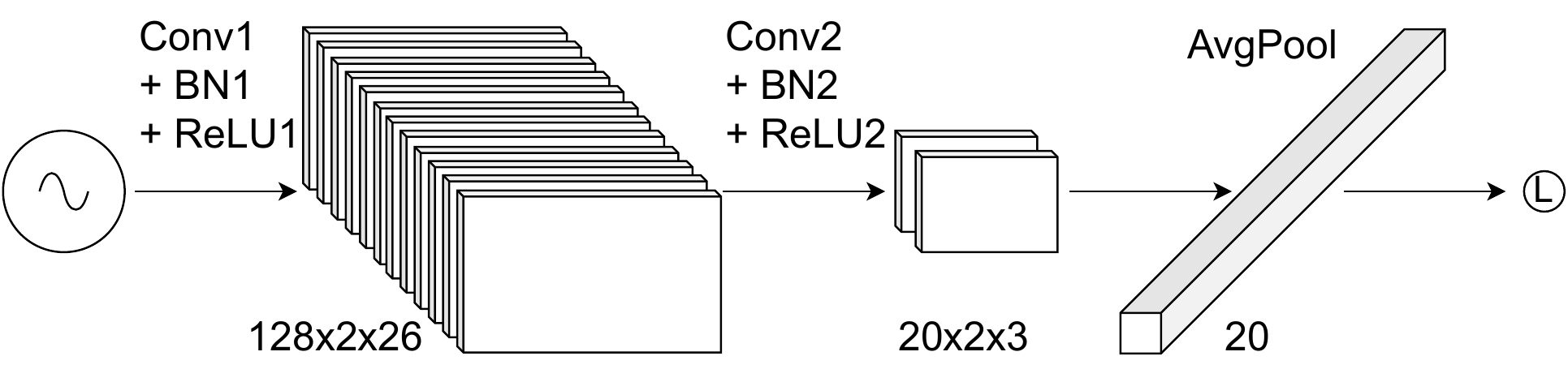}
    \label{fig:rvnn5}} \newline \newline \quad \subfloat{%
    \includegraphics[width=0.45\textwidth]{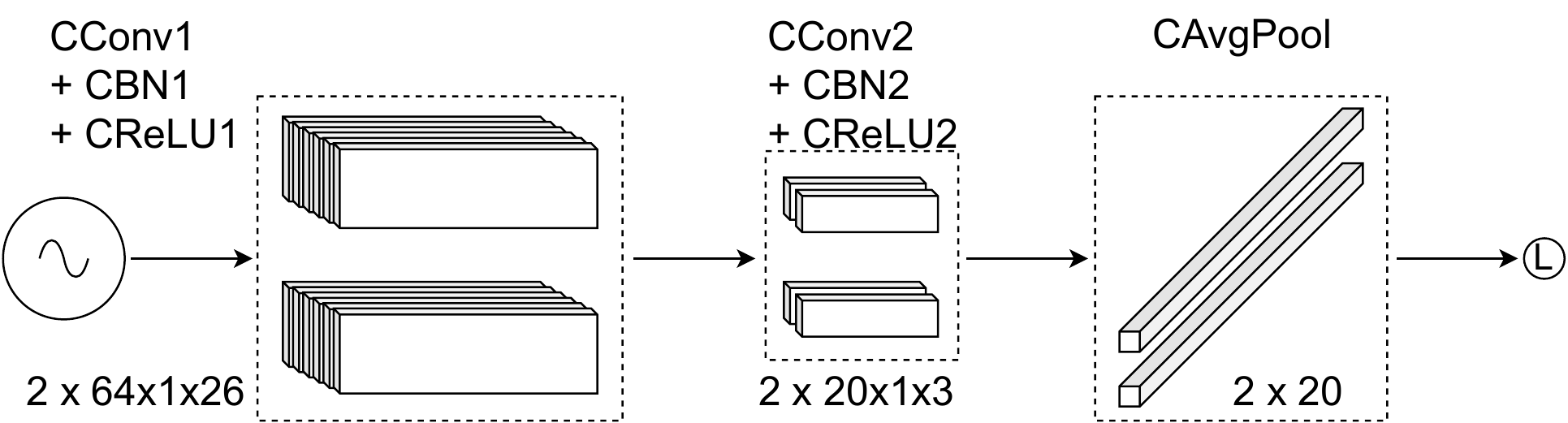}
    \label{fig:cvnn5}}
  \caption{Architecture of RVNN (top) and CVNN (bottom).}
  \label{fig:net_rvnn_cvnn}
\end{figure}

Figure \ref{fig:net_rvnn_cvnn} shows the architecture of the RVNN (top) and
CVNN (bottom) used in our experiments. To simplify the analysis, we used RVNNs and CVNNs
with two convolutional layers; deeper networks would likely improve performance
but at the cost of making the analysis much more complex. Our
baseline RVNN has two real-valued convolutional layers (conv layers), with each
conv layer followed by a batch normalization layer and a ReLU layer. The output
of the second conv layer goes to an average pooling layer, which produces the
output. We use a cross entropy loss function to train the network.

Conceptually, the CVNN has an analogous structure as the RVNN, except each layer
now consists of a real-valued part and a complex-valued part. The details of
both architectures can be found in Table \ref{tab:expr_arch}.
\begin{table}
    \Huge
    \centering \resizebox{0.95\columnwidth}{!}{
    \begin{tabular}{@{}lll|lll@{}}
      \toprule
      \multicolumn{3}{l|}{\thead{RVNN (Real-Valued CNN)}} & \multicolumn{3}{l}{\thead{CVNN (Complex-Valued CNN)}}                               \\
      \toprule
      Layer Type           & Output Size            & Filter/Stride Size & Layer Type & Output Size                    & Filter/Stride Size \\
      \midrule
      Input                & 1$\times$2$\times$100 &                  & Input      & 2$\times$1$\times$1$\times$100 &                         \\
      Conv2D               & 128$\times$2$\times$26 & (1, 25)/(1, 3) & CConv2D    & 2$\times$64$\times$1$\times$26 & (1, 25)/(1, 3)        \\
      BatchNorm            & 128$\times$2$\times$26 &                  & CBatchNorm & 2$\times$64$\times$1$\times$26 &                          \\
      ReLU                 & 128$\times$2$\times$26 &                  & CReLU      & 2$\times$64$\times$1$\times$26 &                          \\
      Conv2D               & 20$\times$2$\times$3   & (1, 20)/(1, 3) & CConv2D    & 2$\times$20$\times$1$\times$3  & (1, 20)/(1, 3)         \\
      BatchNorm            & 20$\times$2$\times$3   &                  & CBatchNorm & 2$\times$20$\times$1$\times$3  &                          \\
      ReLU                 & 20$\times$2$\times$3   &                  & CReLU      & 2$\times$20$\times$1$\times$3  &                          \\
      AvgPool2D            & 20                   & (2, 3)/(1, 1)  & CAvgPool2D & 2$\times$20                   & (1, 3)/(1, 1)          \\
      Cross Entropy of $x$ &                        &                  & Cross Entropy of $|z|$ &                                &              \\
      \bottomrule
    \end{tabular}
  }
\caption{Details of the RVNN and CVNN architectures}
    \label{tab:expr_arch}
\end{table}

To perform a head-to-head comparison between a RVNN and an "equivalent" CVNN, we keep the number of parameters for both types of neural networks equal to each other.
For instance, with the \stthree~datasets, both the RVNN and the CVNN have 54400 parameters total. However, having an equal number of
parameters is not an exact "apples-to-apples" comparison as there are other
architectural differences. As such, we also include experiments that change the input features and ablation experiments to shed light on the capabilities of CVNNs over RVNNs.


\subsection{Ablation Strategies} \label{sec:expr_ab} 
\begin{figure}
  \centering 
  \subfloat {%
    \includegraphics[width=0.45\textwidth]{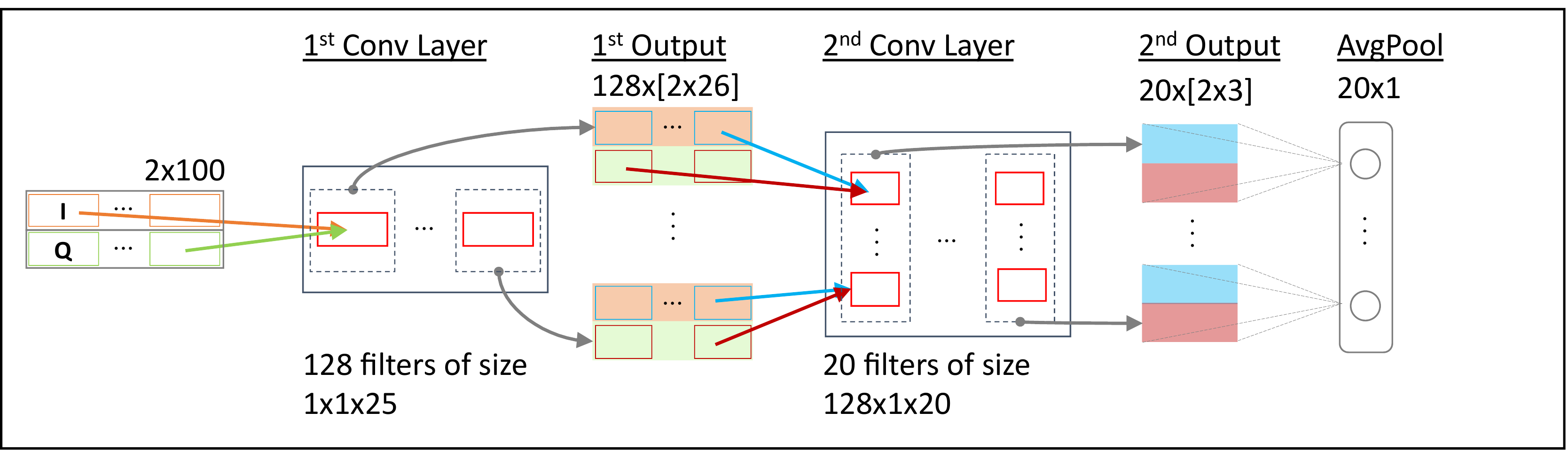}
    \label{fig:cf_rvnn}} \quad 
  \subfloat {%
    \includegraphics[width=0.45\textwidth]{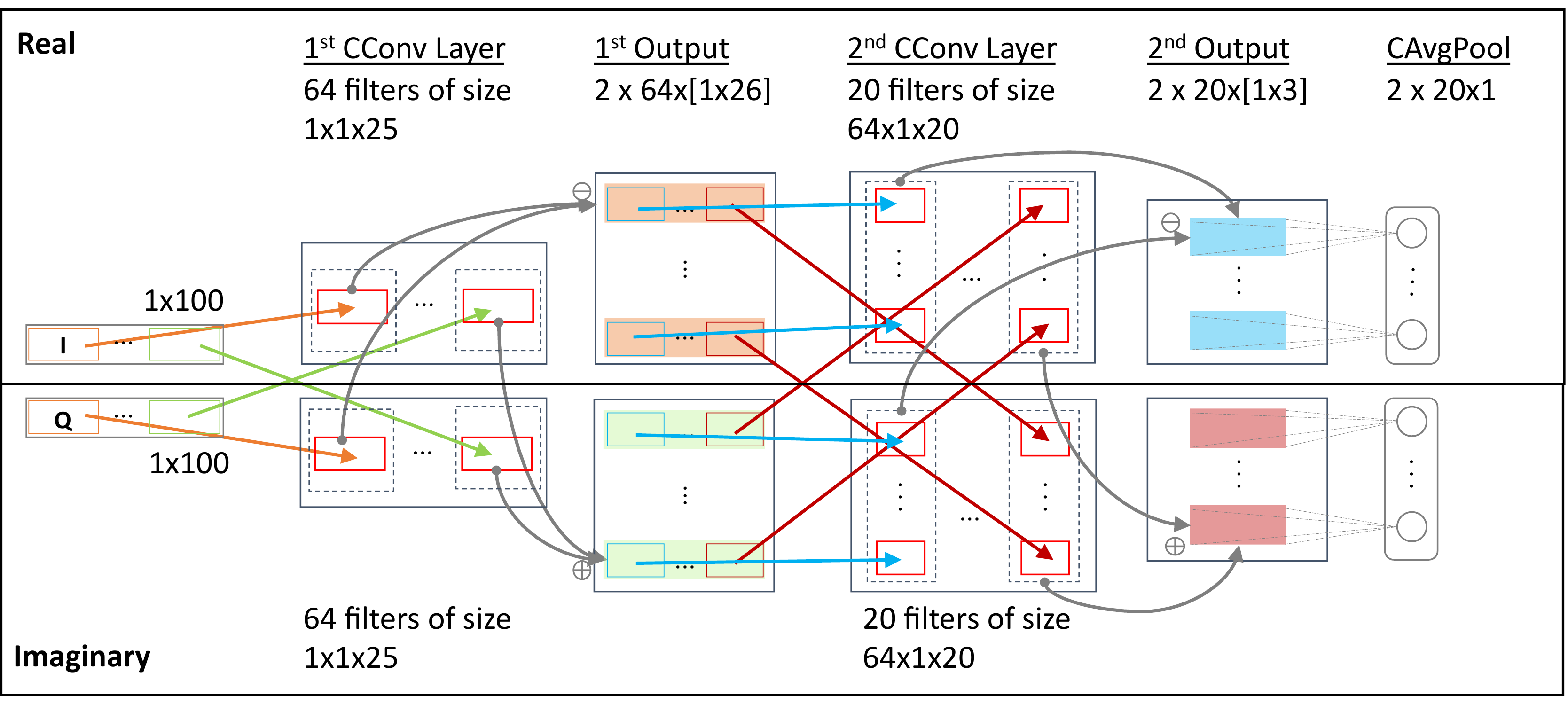}
    \label{fig:cf_cvnn}}
  \caption{The computational flow schematic for a RVNN (top) and a CVNN (bottom).}
  \label{fig:cf_rvnn_cvnn}
\end{figure}

The main difference between
the CVNN and the RVNN is due to the cross terms between the real and imaginary
parts in the multiplication and convolution operations. Figure \ref{fig:cf_rvnn_cvnn} illustrates the computational flow for a
RVNN (top) and a CVNN (bottom); the computational flow of a CVNN is much more complex than that
of a RVNN. As a concrete example, take the first complex-valued convolutional
layer ($1^{st}$ CConv Layer) and its associated Output ($1^{st}$ Output) in
Figure \ref{fig:cf_cvnn}. The real part of the $1^{st}$ Output (orange blocks)
comes from the linear combination of I data scanned by real filters and Q data
scanned by imaginary filters (as shown by the grey arrows pointing to the
$\ominus$), while the imaginary part of the $1^{st}$ Output (green blocks) comes
from the cross terms formed by the linear combination of I data scanned by
imaginary filters and Q data scanned by real filters (as shown by the gray
arrows pointing to $\oplus$). Here the $\ominus$ and $\oplus$ symbols in Figure
\ref{fig:cf_cvnn} correspond to the real ($\mathbf{A} * \mathbf{x}-\mathbf{B} *
\mathbf{y}$) and imaginary parts ($\mathbf{B} * \mathbf{x}+\mathbf{A} *
\mathbf{y}$) of Equation \ref{eq:ccov1}.


 
We perform a deeper analysis into the differences between the two models through a
series of ablation experiments which remove parts of the CVNN, specifically
parts that correspond to some of the cross terms. By "remove" we mean that we
zero out the ablated component. If performance degrades significantly, then the
portion of the CVNN that is removed is an important component of the model.
Figure \ref{fig:cf_cvnn} illustrates the CVNN as 2 layers, with each layer
consisting of convolutional filters and an output. There is a set of 12 ablation
experiments that can be defined by the following dimensions and values: 1) the
layer to ablate: $1^{st}$ layer ({\bf L1}), $2^{nd}$ layer ({\bf L2}), or both
({\bf L12}); 2) the component of the layer to ablate: the convolutional filter
({\bf C}) or the output ({\bf O}); 3) the part of the component to remove: the
real part ({\bf RE}) or the imaginary part ({\bf IM}).



Table \ref{tab:expr_abs} summarizes the 12 ablation experiments, with the name
of the experiment stated in each table cell. The naming convention follows the
boldface labels chosen for each ablation dimension, concatenated with an
underscore. For example, "L1\_O\_IM" ablates the $1^{st}$ Output and remove the
imaginary part of the output (green blocks of the $1^{st}$ Output in Figure
\ref{fig:cf_cvnn}). In our results, we only show results of the "IM" ablations
which remove the imaginary part (of the output or convolutional filter) and keep
the real part; the results of the "RE" ablations have equivalent performance.

\begin{table}
  \centering \resizebox{0.95\columnwidth}{!}{
    \begin{tabular}{@{}l|ll|ll@{}}
      \toprule
      \multicolumn{1}{l|}{\thead{}} & \multicolumn{2}{l|}{\thead{Ablate Outputs}} & \multicolumn{2}{l}{\thead{Ablate Conv Filters}} \\
      \toprule
      & Remove Re Output & Remove Im Output & Remove Re Conv   & Remove Im Conv \\
      \midrule
      L1 (Ablate $1^{st}$ Layer) & L1\_O\_RE  & \textbf{L1\_O\_IM}  & L1\_C\_RE & \textbf{L1\_C\_IM}                                                  \\
      L2 (Ablate $2^{nd}$ Layer) & L2\_O\_RE  & \textbf{L2\_O\_IM}  & L2\_C\_RE & \textbf{L2\_C\_IM}                                                  \\
      L12 (Ablate Both)          & L12\_O\_RE & \textbf{L12\_O\_IM} & L12\_C\_RE & \textbf{L12\_C\_IM}                                                \\
      \bottomrule
    \end{tabular}
  }
  \caption{Ablation Models}
  \label{tab:expr_abs}
\end{table}

\subsection{Varying the Inputs to the Neural Networks}
We also explore changing the set of inputs to the neural networks. For IQ
samples, we investigate the following three configurations: 1) use only the
input {\bf I}; 2) use only the input {\bf Q}; 3) use both I and Q ({\bf IQ}
for short) as the input.
For the polar representation of IQ samples, we denote the magnitude part as R
and the phase part as T. As before, we explore three
configurations: 1) use only the input {\bf R}; 2) use only the input {\bf
  T}; 3) use both R and T ({\bf RT} for short).
For configurations that only provide one input, we zero out the other input.



\section{Performance Results and Analysis}
\label{sec:expr}

In this section, we report the classification results and present the findings on the comparison between RVNN and CVNN and between the various ablation models of CVNN, using the datasets described in
Section~\ref{sec:data}. In addition, we investigate how the different input modes affect the performance. We present the average accuracy of each model for
each dataset scenario as a bar chart, 
error bars ($\pm\sigma$) are also shown. 

\subsection{RVNNs versus CVNNs}

\begin{figure}
  \centering 
  \subfloat[\dsone]{%
    \includegraphics[width=0.13\textwidth]{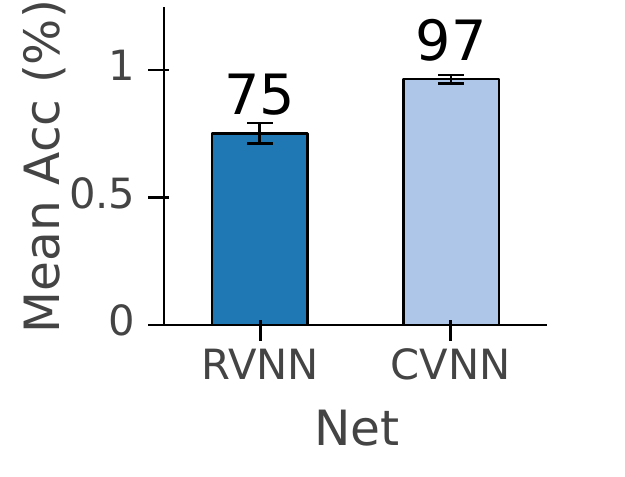}
    \label{fig:a1_ost1}} \hspace*{-1em}
  \subfloat[\dstwo]{%
    \includegraphics[width=0.13\textwidth]{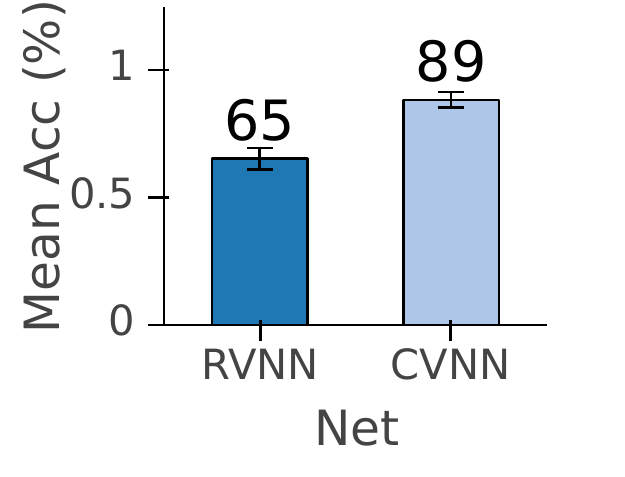}
    \label{fig:a1_ost2}} \hspace*{-1em}
  \subfloat[\stthree]{%
    \includegraphics[width=0.13\textwidth]{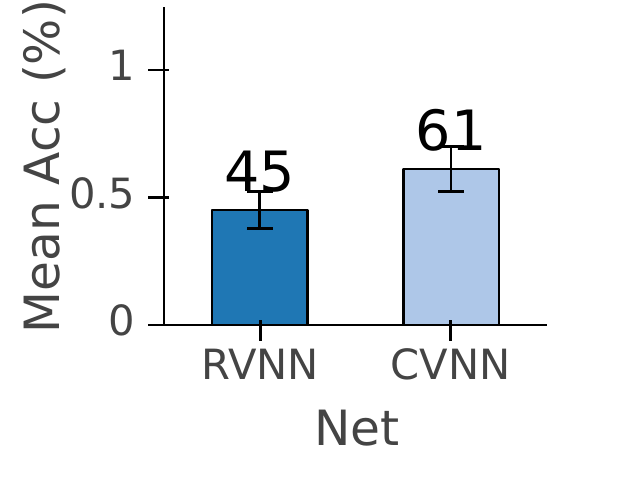}
    \label{fig:a1_st3}} \hspace*{-1em}
  \subfloat[\stfour]{%
    \includegraphics[width=0.13\textwidth]{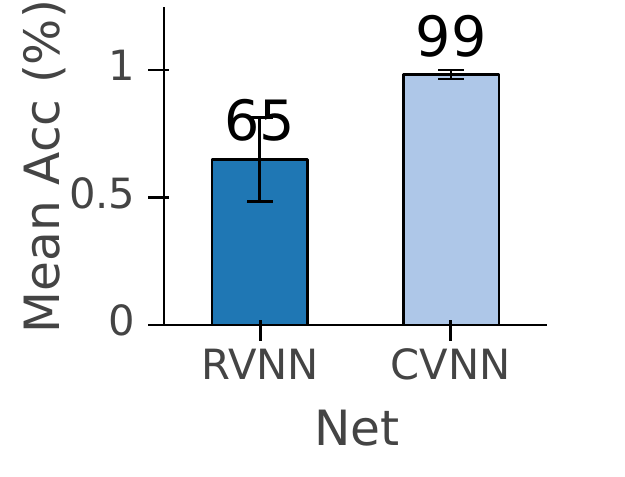}
    \label{fig:a1_st4}}
  %
  \caption{RVNN vs. CVNN (Input: IQ)}
  \label{fig:expr_sec_a1}
\end{figure}

{\bf Using IQ Data as Input:} Figure \ref{fig:expr_sec_a1} shows the average
classification accuracy achieved using RVNNs and CVNNs with raw IQ samples as
input under each of the four dataset scenarios (\dsone, \dstwo, \stthree, \stfour); CVNNs outperform RVNNs by a substantial amount (16\%-34\%) for
all four scenarios. 
In the \dsosu~case, the \dsone~has better performance than
\dstwo, which is expected since Indoor signals have lesser interference compared
with the Outdoor ones. For the \dsne~case, interestingly, our experimental
results reveal that \stfour~(Anechoic Chamber WiFi) achieves higher accuracy
than that achieved under \stthree~(Wired) for both RVNN and CVNN.

\begin{figure}
  \centering 
  \subfloat[\dsone]{%
    \includegraphics[width=0.13\textwidth]{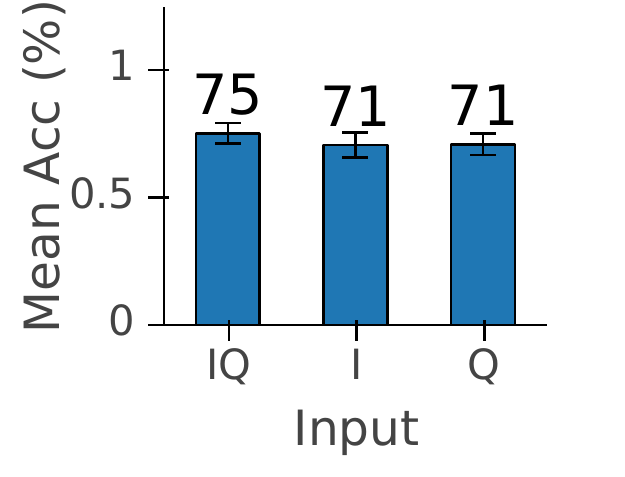}
    \label{fig:a2_ost1}} \hspace*{-1em} 
  \subfloat[\dstwo]{%
    \includegraphics[width=0.13\textwidth]{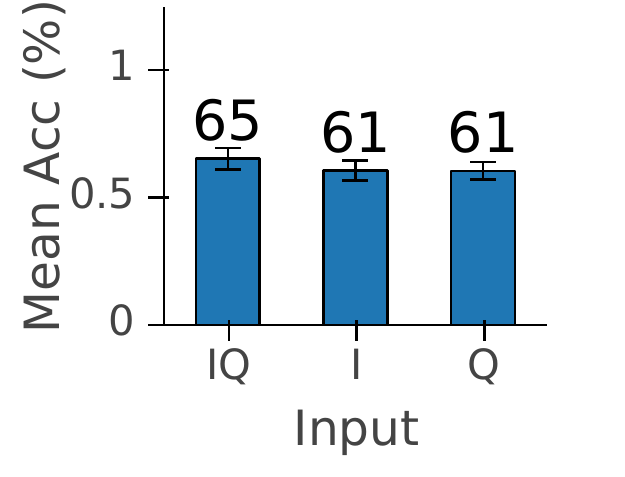}
    \label{fig:a2_ost2}} \hspace*{-1em} 
  \subfloat[\stthree]{%
    \includegraphics[width=0.13\textwidth]{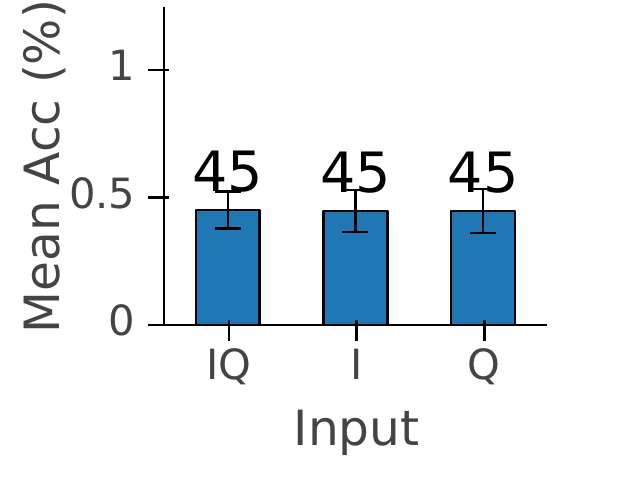}
    \label{fig:a2_st3}} \hspace*{-1em} 
  \subfloat[\stfour]{%
    \includegraphics[width=0.13\textwidth]{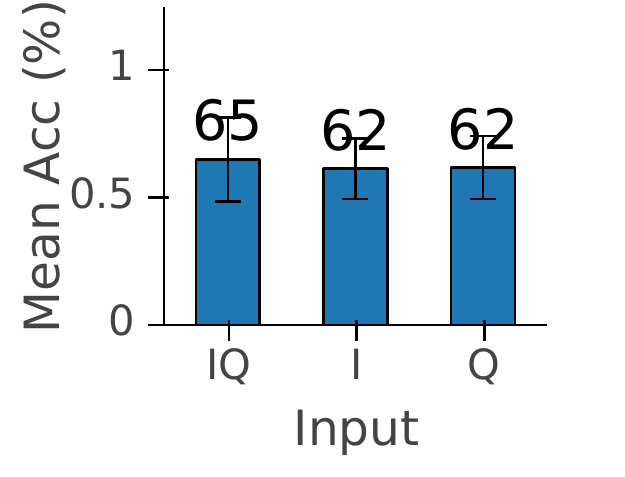}
    \label{fig:a2_st4}}
  %
\newline

  \subfloat[\dsone]{%
    \includegraphics[width=0.13\textwidth]{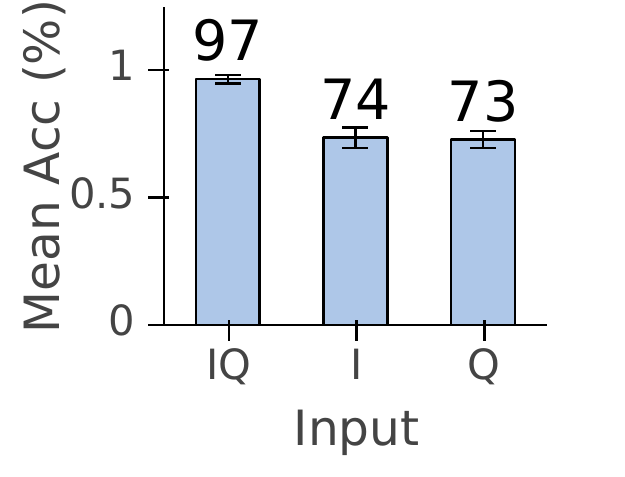}
    \label{fig:a3_ost1}} \hspace*{-1em}
  \subfloat[\dstwo]{%
    \includegraphics[width=0.13\textwidth]{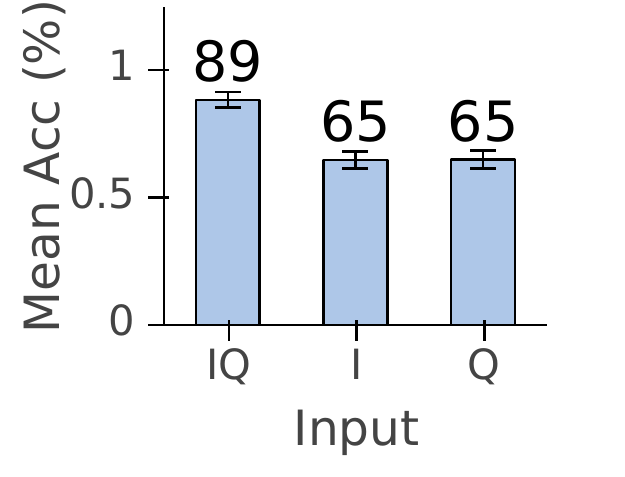}
    \label{fig:a3_ost2}} \hspace*{-1em} 
  \subfloat[\stthree]{%
    \includegraphics[width=0.13\textwidth]{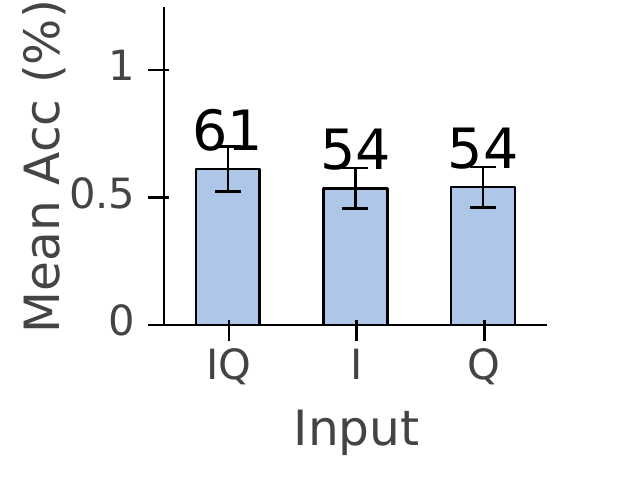}
    \label{fig:a3_st3}} \hspace*{-1em}
  \subfloat[\stfour]{%
    \includegraphics[width=0.13\textwidth]{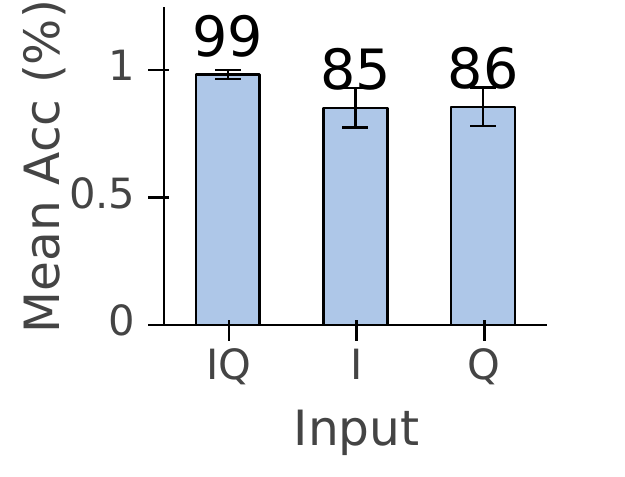}
    \label{fig:a3_st4}}
  %
  \caption{IQ vs. I vs. Q. (Top: RVNN, Bottom: CVNN)}
  \label{fig:expr_sec_a3}
\end{figure}

{\bf Using I vs Q vs IQ Data as Input:}
%
Figure \ref{fig:expr_sec_a3} shows the average
accuracy for the RVNN (top) and CVNN (bottom) models, when considering I only, Q
only, and both IQ as data input to the learning models. The figures show that the I and Q components when fed by themselves as an input to the learning model
yield roughly the same average accuracy, and this is true for each of four
studied datasets and each of the two models (RVNN or CVNN). This result
indicates that in isolation, I and Q contain a similar amount of predictive
information. Note that using both IQ as input results in a (sometimes slightly)
higher accuracy, which
is expected since the full input should contain more information.
However, the increase in accuracy from using both IQ over using I or Q in
isolation is much higher for CVNNs (7\%-24\%) than RVNNs ($<$5\%). This result
shows that a CVNN is able to use the joint information between I and Q more
effectively than a RVNN due to the cross terms generated by the complex number
multiplication and convolution operations.


\begin{figure}[t!]
  \centering 
  \subfloat[\dsone]{%
    \includegraphics[width=0.13\textwidth]{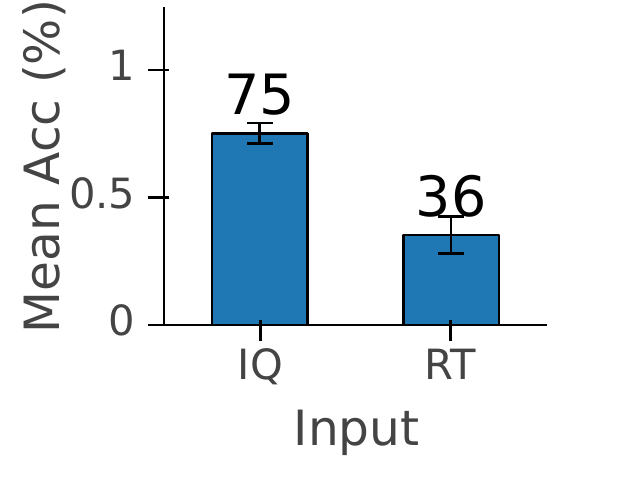}
    \label{fig:b1_ost1}} \hspace*{-1em}
  \subfloat[\dstwo]{%
    \includegraphics[width=0.13\textwidth]{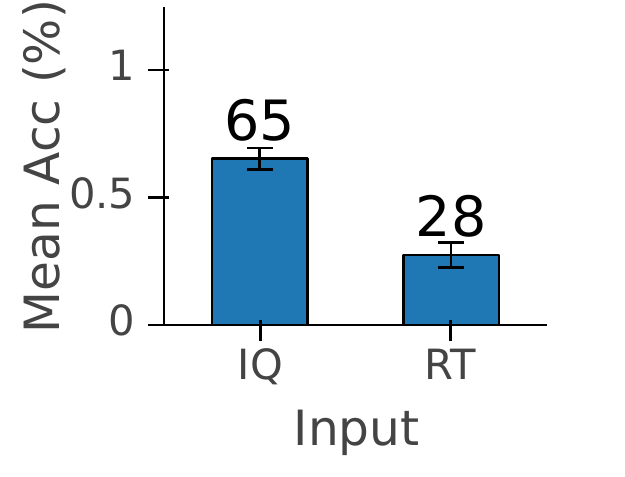}
    \label{fig:b1_ost2}} \hspace*{-1em} 
  \subfloat[\stthree]{%
    \includegraphics[width=0.13\textwidth]{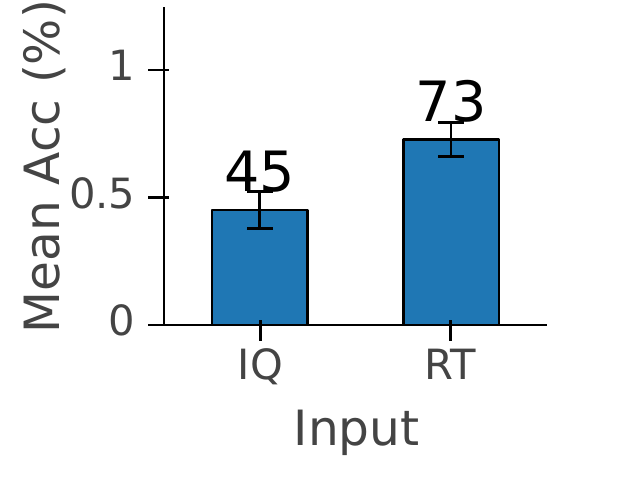}
    \label{fig:b1_st3}} \hspace*{-1em}
  \subfloat[\stfour]{%
    \includegraphics[width=0.13\textwidth]{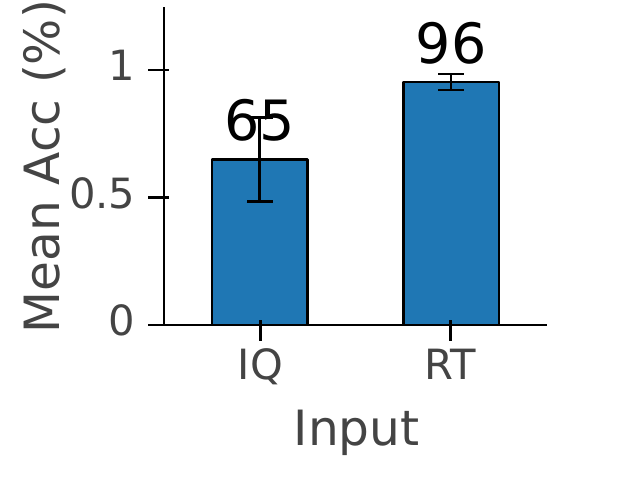}
    \label{fig:b1_st4}}
    \newline
  %

  \subfloat[\dsone]{%
    \includegraphics[width=0.13\textwidth]{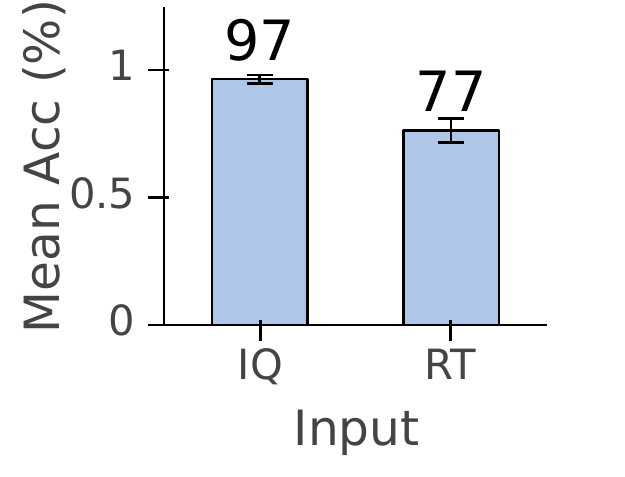}
    \label{fig:b2_ost1}} \hspace*{-1em} 
  \subfloat[\dstwo]{%
    \includegraphics[width=0.13\textwidth]{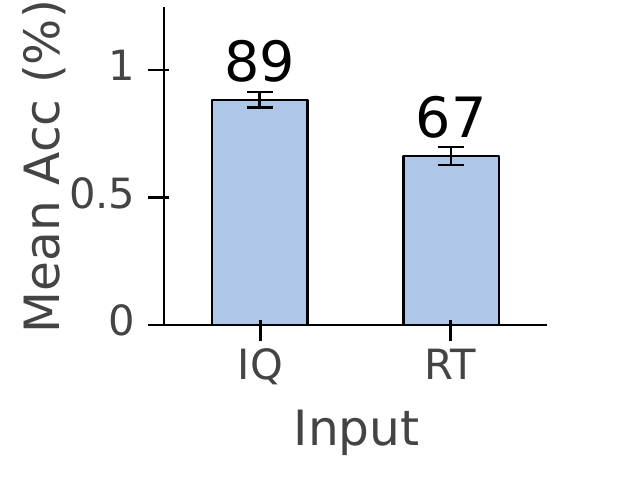}
    \label{fig:b2_ost2}} \hspace*{-1em} 
  \subfloat[\stthree]{%
    \includegraphics[width=0.13\textwidth]{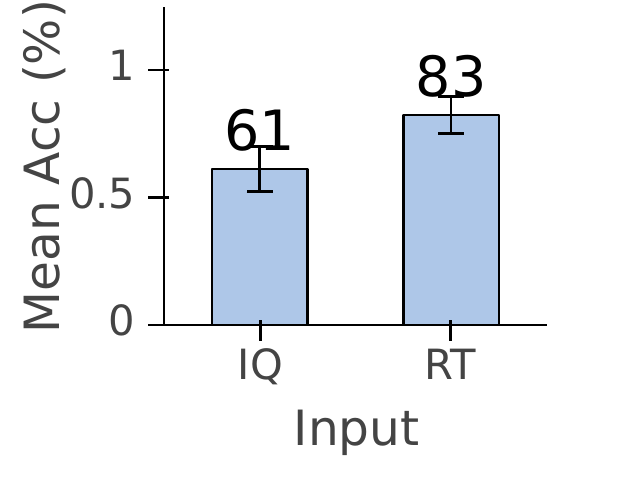}
    \label{fig:b2_st3}} \hspace*{-1em} 
  \subfloat[\stfour]{%
    \includegraphics[width=0.13\textwidth]{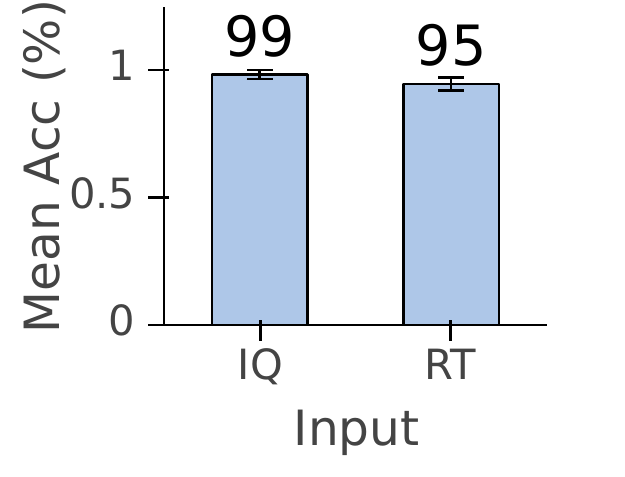}
    \label{fig:b2_st4}}
  %
  \caption{IQ vs. RT. (Top: RVNN, Bottom: CVNN)}
  \label{fig:expr_sec_b2}
\end{figure}

{\bf Using RT vs IQ Data as Input: } Figure \ref{fig:expr_sec_b2} (top) shows that
the IQ representation of the input data produces a large improvement over RT
(approximately 40\%) for the \dsosu~datasets under RVNNs. The opposite is true
for the \dsne~datasets as the RT representation produces improvements of 28\%-31\% over the IQ representation. Figure \ref{fig:expr_sec_b2} (bottom) shows that for
CVNNs, the IQ representation produces improvements over the RT representation
(4\%-22\%) in three datasets, with the exception being for \stthree, which
resulted in the IQ representation being 22\% lower. When comparing RT results of
CVNNs to those of RVNNs on specific datasets, we can see that CVNNs produce an
accuracy that is much higher (10\%-41\%) than that produced under RVNNs in all
datasets except for \stfour, which yields close to equivalent accuracy.


\begin{figure}[t!]
  \centering 
  \subfloat[\dsone]{%
    \includegraphics[width=0.13\textwidth]{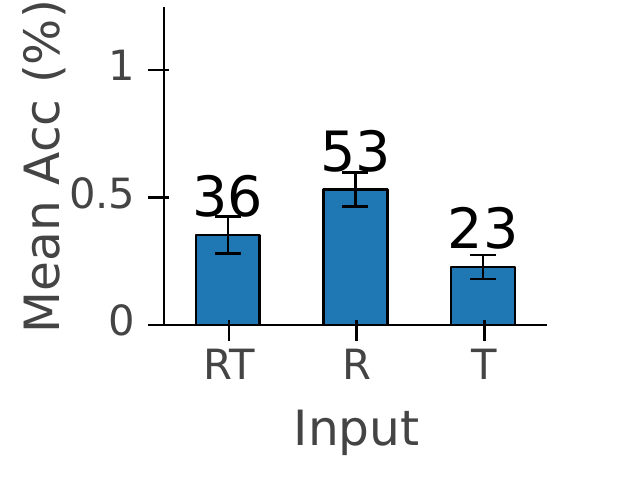}
    \label{fig:b3_ost1}} \hspace*{-1em} 
  \subfloat[\dstwo]{%
    \includegraphics[width=0.13\textwidth]{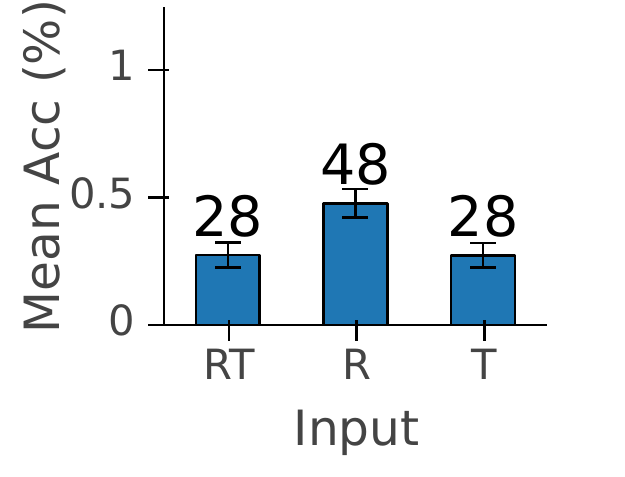}
    \label{fig:b3_ost2}} \hspace*{-1em} 
  \subfloat[\stthree]{%
    \includegraphics[width=0.13\textwidth]{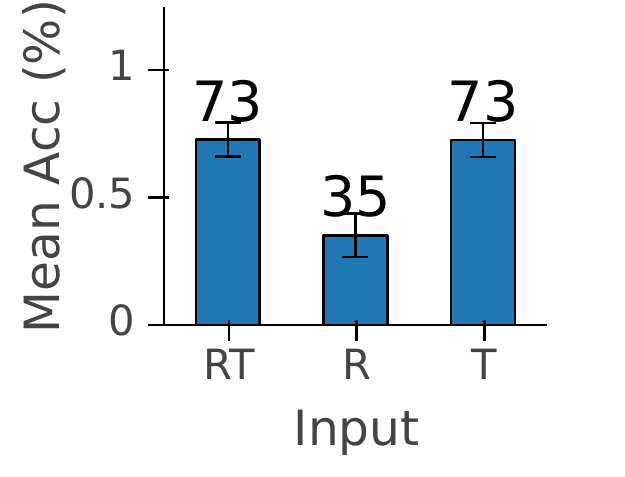}
    \label{fig:b3_st3}} \hspace*{-1em}
  \subfloat[\stfour]{%
    \includegraphics[width=0.13\textwidth]{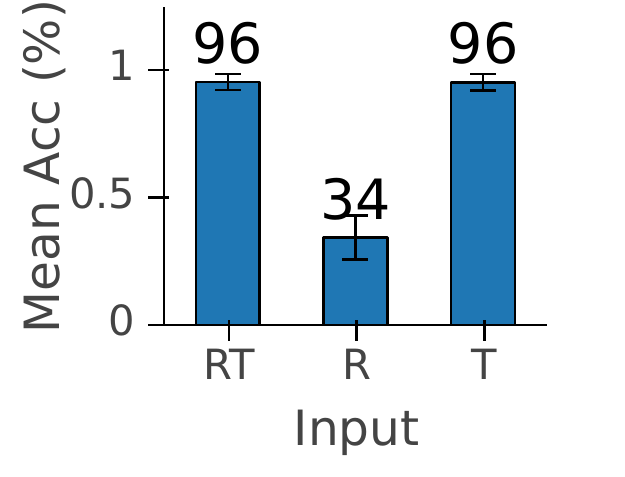}
    \label{fig:b3_st4}}
    \newline
  %

  \subfloat[\dsone]{%
    \includegraphics[width=0.13\textwidth]{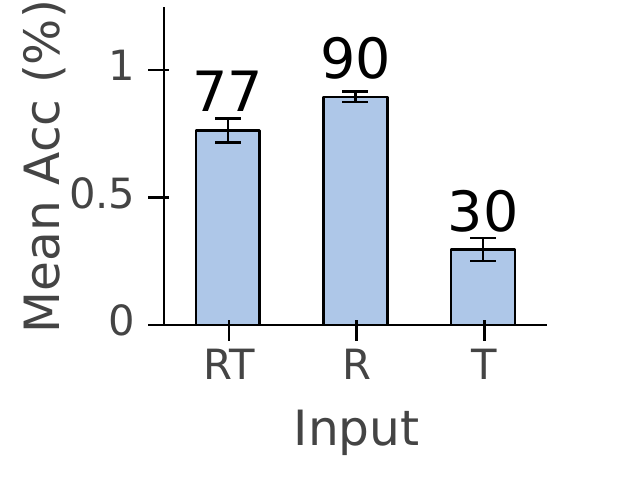}
    \label{fig:b4_ost1}} \hspace*{-1em} 
  \subfloat[\dstwo]{%
    \includegraphics[width=0.13\textwidth]{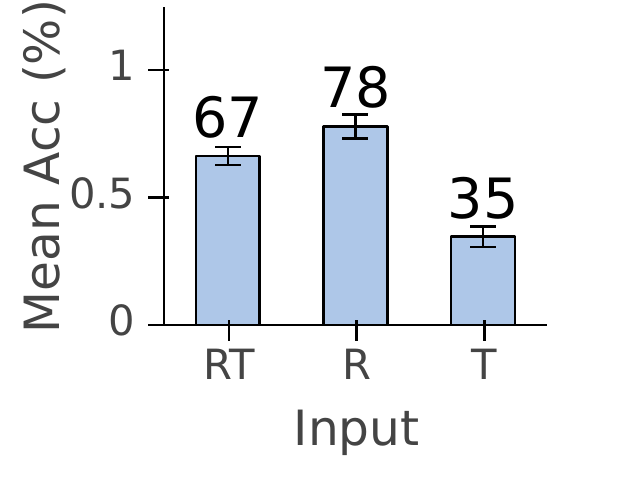}
    \label{fig:b4_ost2}} \hspace*{-1em} 
  \subfloat[\stthree]{%
    \includegraphics[width=0.13\textwidth]{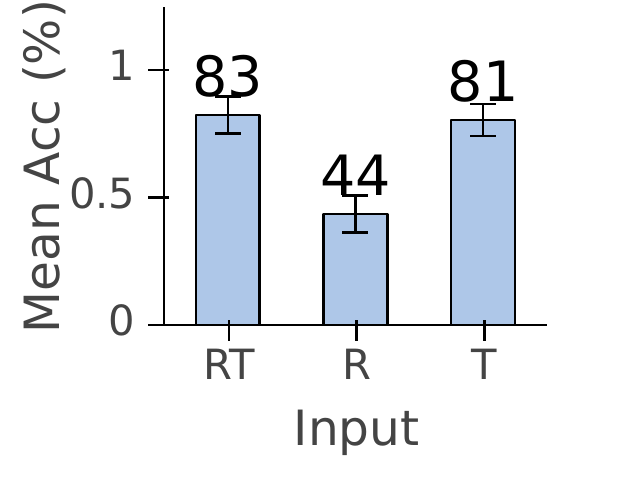}
    \label{fig:b4_st3}} \hspace*{-1em}
  \subfloat[\stfour]{%
    \includegraphics[width=0.13\textwidth]{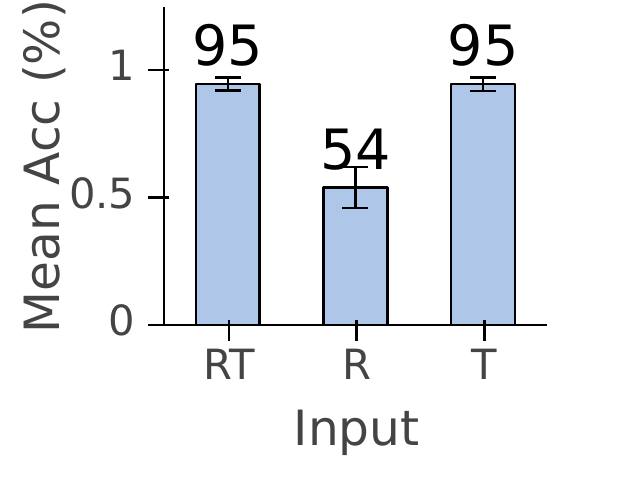}
    \label{fig:b4_st4}}
  %
  \caption{RT vs. R vs. T (Top: RVNN, Bottom: CVNN)}
  \label{fig:expr_sec_b4}
\end{figure}

Figure \ref{fig:expr_sec_b4} shows that for
\dsosu~datasets, having R as the only input produces large improvements in
accuracy of 20\%-60\% compared to having only T as input and improvements of
11\%-20\% over having both RT as inputs. For \dsne~datasets, the results are the
opposite. Having T as the only input produces improvements in accuracy of
37\%-62\% than having only R. However, having both RT as inputs is
almost equivalent to the accuracy of only including T. In general, CVNNs tend to
outperform the corresponding R, T or RT results on RVNNs with the exception of
the RT and T results for the \stfour~dataset.

\subsection{Ablation Analysis} 

\begin{figure}
  \centering 
  \subfloat[\dsone]{%
    \includegraphics[width=0.13\textwidth]{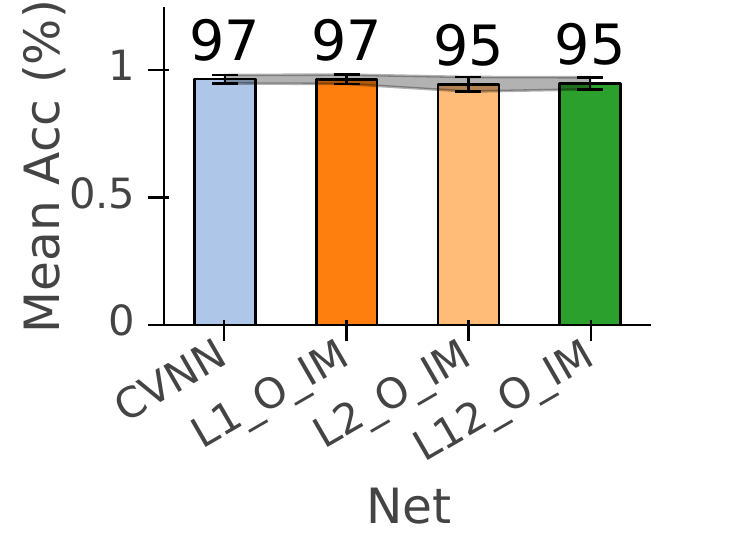}
    \label{fig:c1_ost1}} \hspace*{-1em}
  \subfloat[\dstwo]{%
    \includegraphics[width=0.13\textwidth]{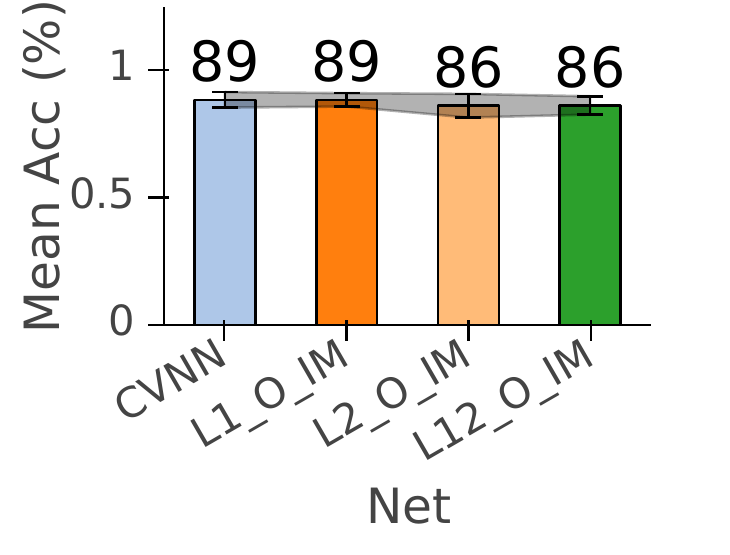}
    \label{fig:c1_ost2}} \hspace*{-1em} 
  \subfloat[\stthree]{%
    \includegraphics[width=0.13\textwidth]{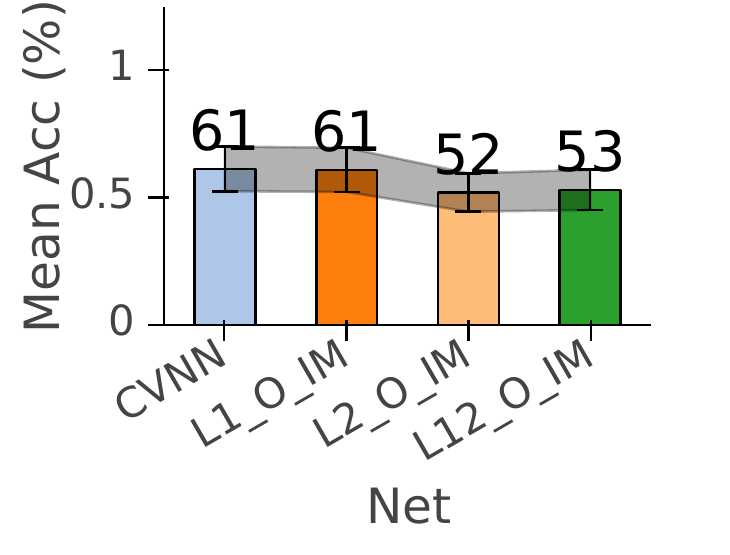}
    \label{fig:c1_st3}} \hspace*{-1em}
  \subfloat[\stfour]{%
    \includegraphics[width=0.13\textwidth]{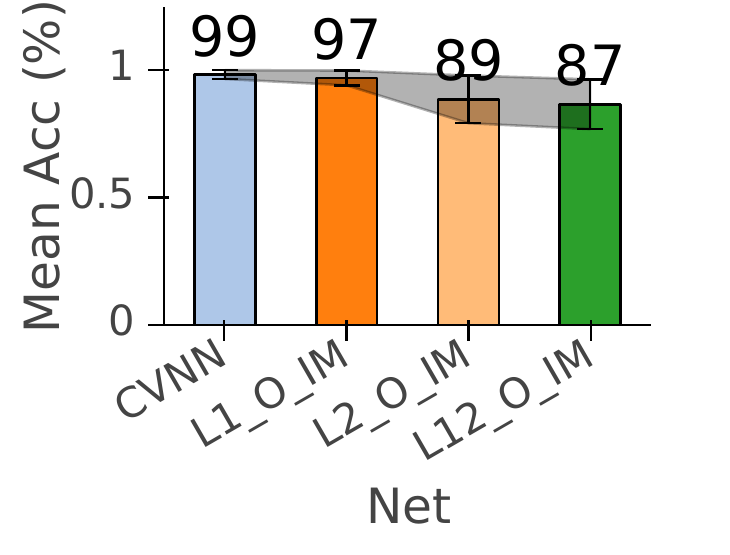}
    \label{fig:c1_st4}} \quad 
  \subfloat[\dsone]{%
    \includegraphics[width=0.13\textwidth]{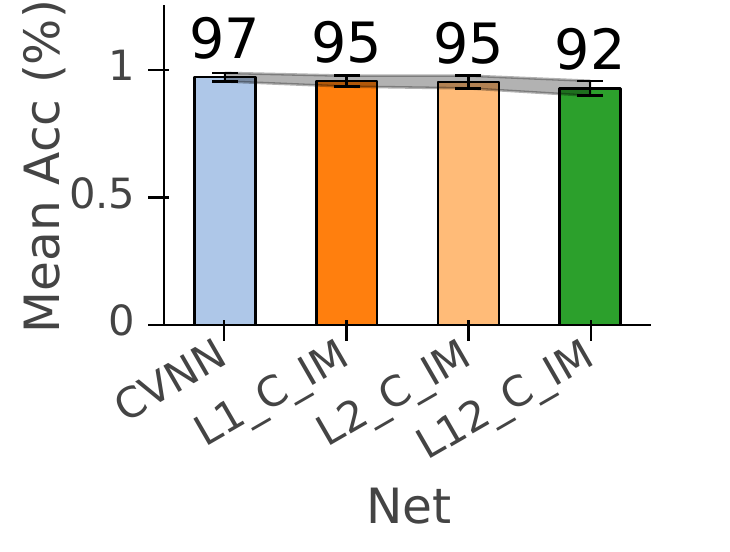}
    \label{fig:c3_ost1}} \hspace*{-1em} 
  \subfloat[\dstwo]{%
    \includegraphics[width=0.13\textwidth]{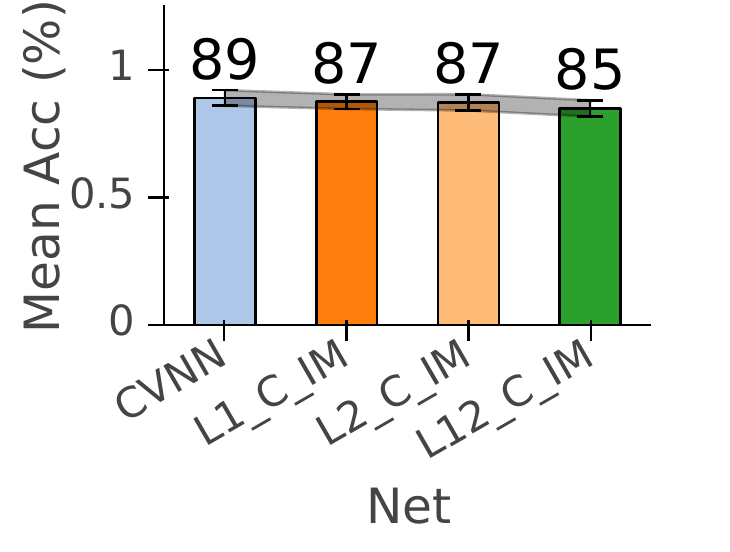}
    \label{fig:c3_ost2}} \hspace*{-1em} 
  \subfloat[\stthree]{%
    \includegraphics[width=0.13\textwidth]{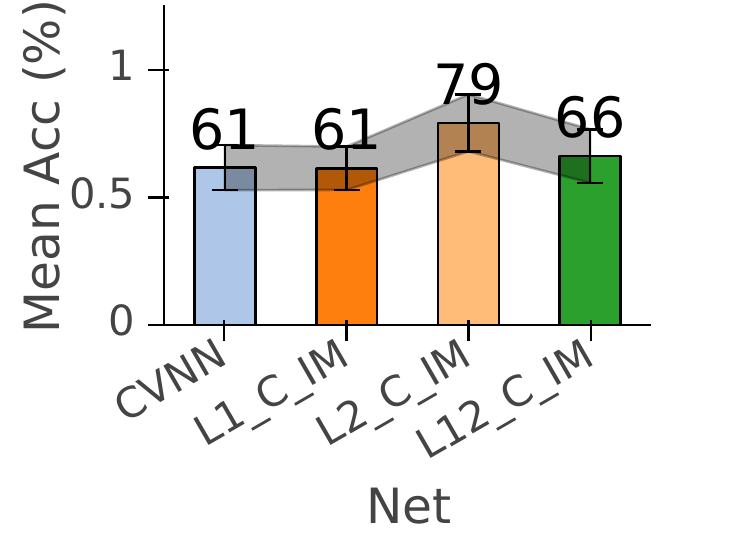}
    \label{fig:c3_st3}} \hspace*{-1em}
  \subfloat[\stfour]{%
    \includegraphics[width=0.13\textwidth]{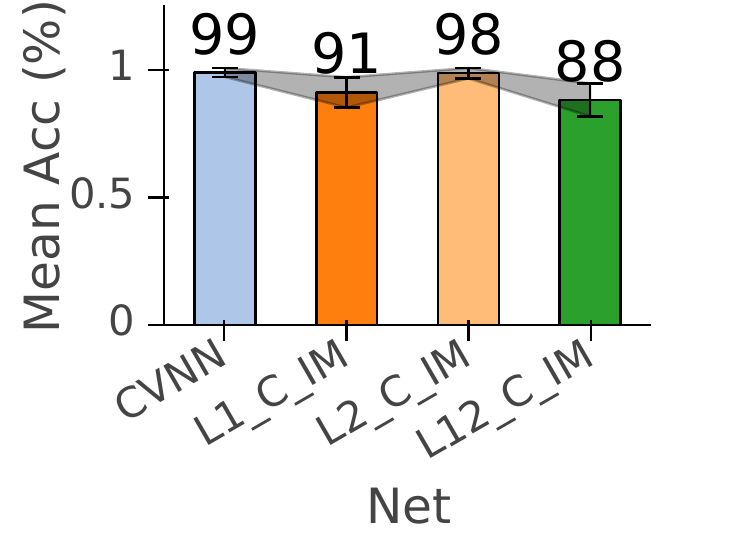}
    \label{fig:c3_st4}}
  \caption{Ablation analysis (Input: IQ). The top row ablates the outputs while
    the bottom row ablates the conv layers}
  \label{fig:expr_sec_c13}
\end{figure}

\begin{figure}
  \centering 
  \subfloat[\dsone]{%
    \includegraphics[width=0.13\textwidth]{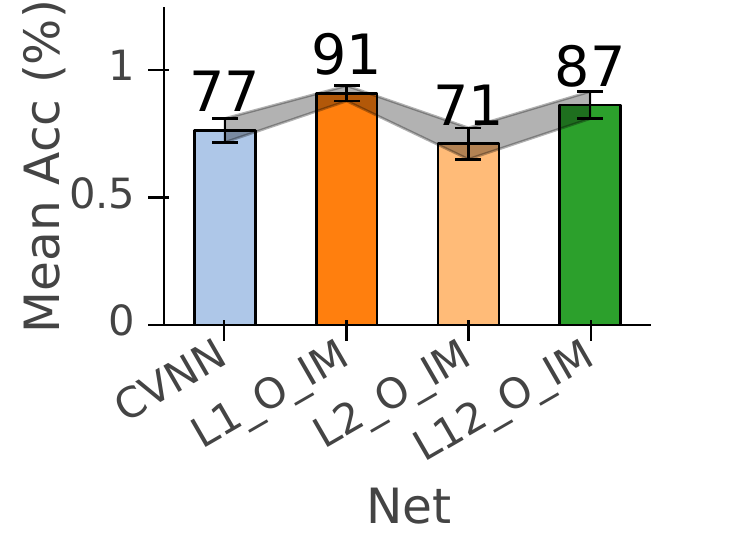}
    \label{fig:c2_ost1}} \hspace*{-1em} 
  \subfloat[\dstwo]{%
    \includegraphics[width=0.13\textwidth]{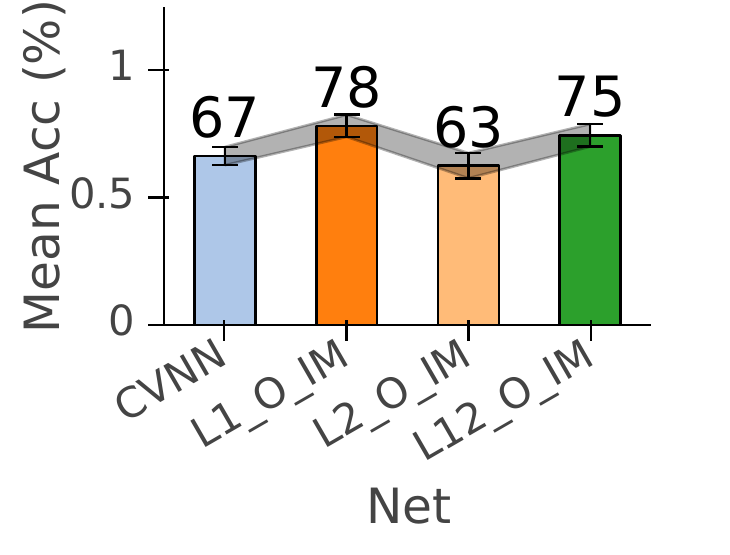}
    \label{fig:c2_ost2}} \hspace*{-1em}
  \subfloat[\stthree]{%
    \includegraphics[width=0.13\textwidth]{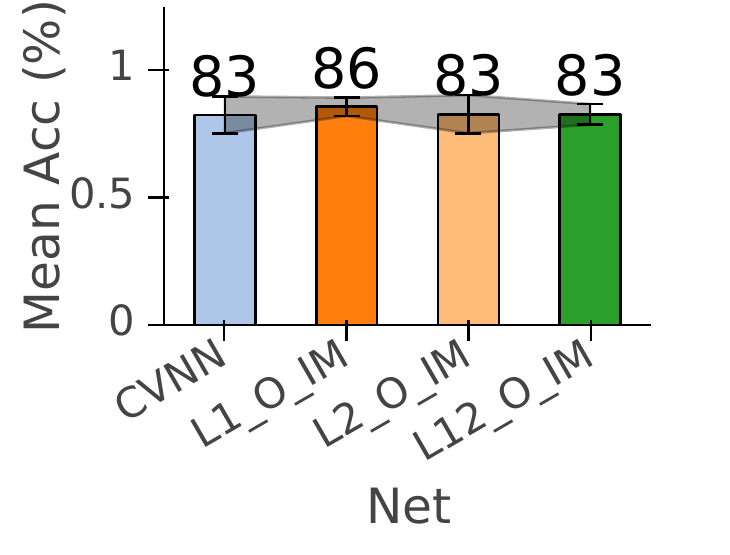}
    \label{fig:c2_st3}} \hspace*{-1em}
  \subfloat[\stfour]{%
    \includegraphics[width=0.13\textwidth]{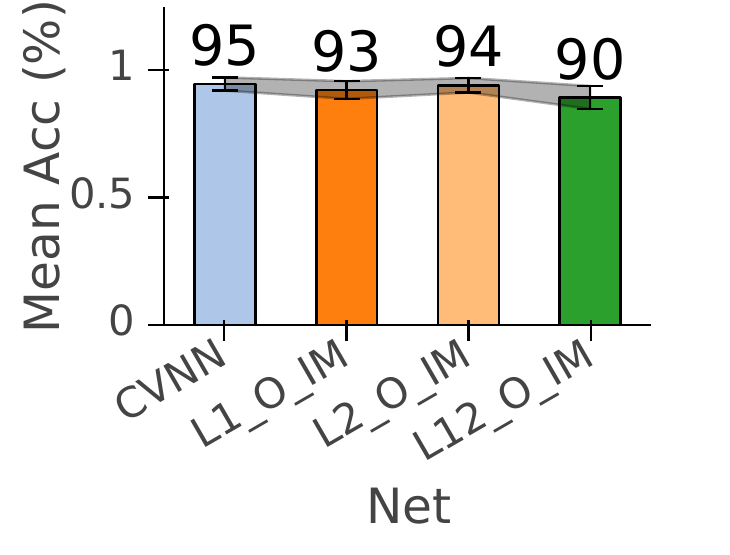}
    \label{fig:c2_st4}} \quad
  \subfloat[\dsone]{%
    \includegraphics[width=0.13\textwidth]{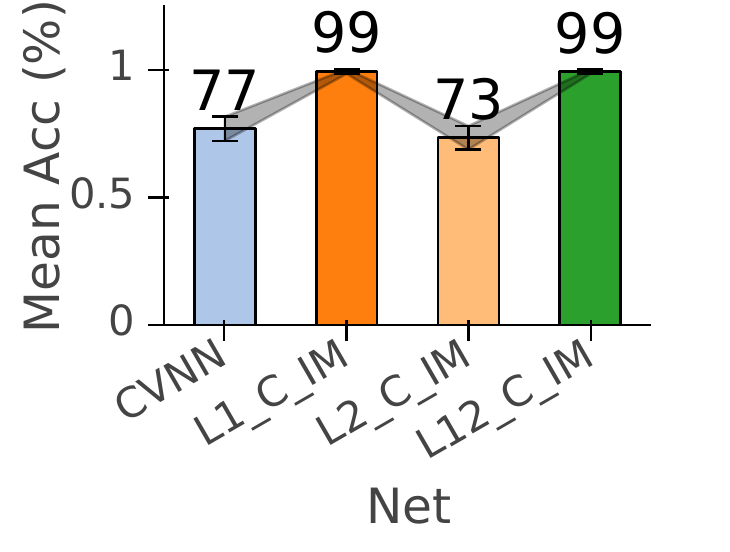}
    \label{fig:c4_ost1}} \hspace*{-1em}
  \subfloat[\dstwo]{%
    \includegraphics[width=0.13\textwidth]{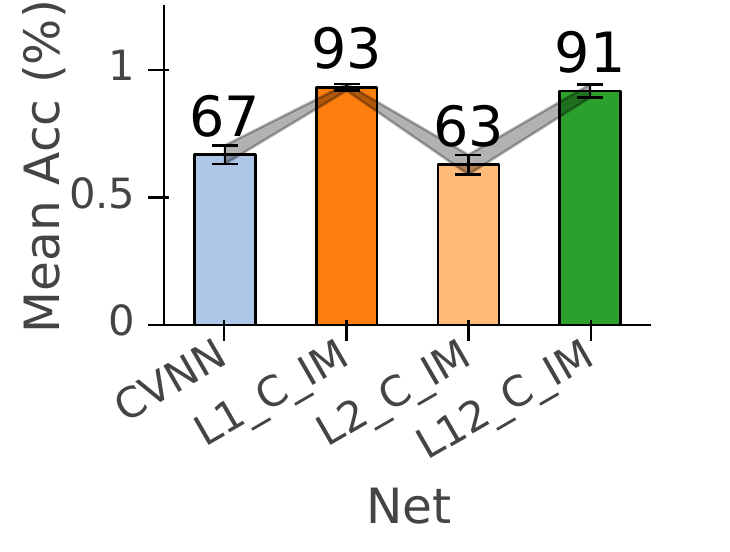}
    \label{fig:c4_ost2}} \hspace*{-1em}
  \subfloat[\stthree]{%
    \includegraphics[width=0.13\textwidth]{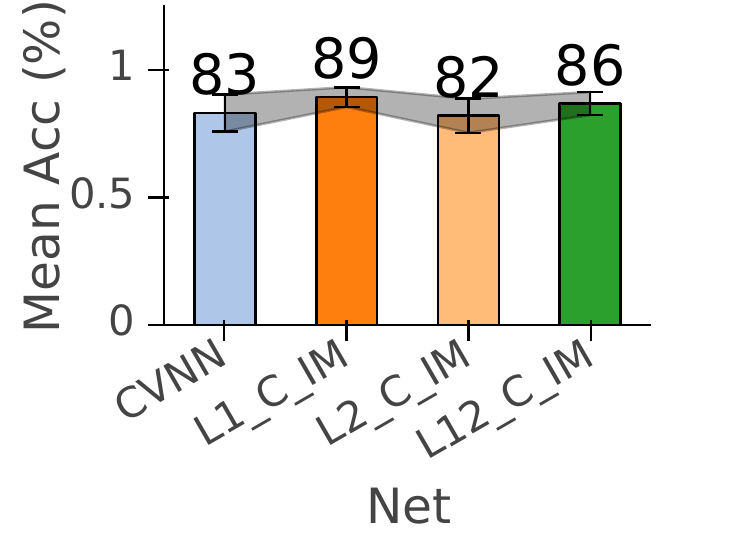}
    \label{fig:c4_st3}} \hspace*{-1em}
  \subfloat[\stfour]{%
    \includegraphics[width=0.13\textwidth]{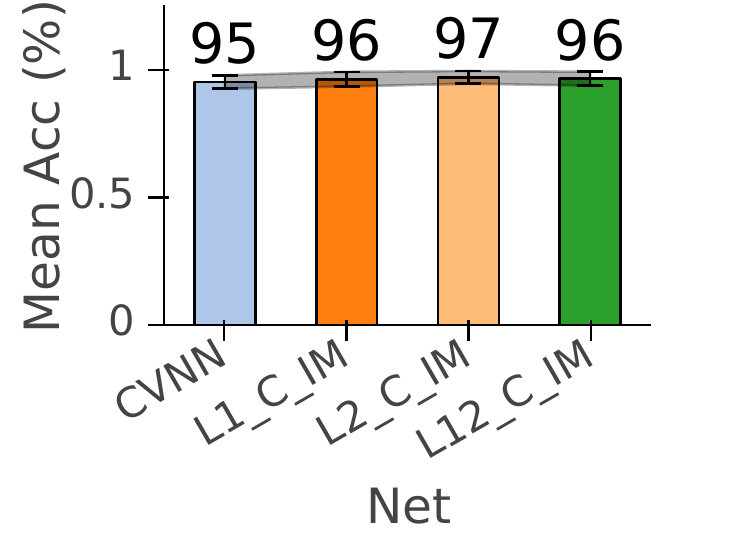}
    \label{fig:c4_st4}}
  \caption{Ablation analysis (Input: RT). The top row ablates the outputs while
    the bottom row ablates the conv layers}
  \label{fig:expr_sec_c24}
\end{figure}

Figure \ref{fig:expr_sec_c13} shows the effect on accuracy from ablating
components of a CVNN (in the order of L1, L2, L12) with IQ as inputs. The first
row (Figures \ref{fig:c1_ost1}--\ref{fig:c1_st4}) shows the results of ablating
the imaginary outputs and the second row (\ref{fig:c3_ost1}--\ref{fig:c3_st4})
shows the results of ablating the imaginary convolutional filters.
Figure \ref{fig:expr_sec_c24} illustrates the results of the analogous
experiments with RT as input.


In general, ablating the outputs produces a decreasing trend in performance
(moving from L1 to L2 to L12) as shown in Figures \ref{fig:expr_sec_c13} and
\ref{fig:expr_sec_c24}. In multi-layer neural networks, deeper layers are known
to represent higher-level features (e.g. parts and objects) rather than
low-level features (e.g. texture and colors) \cite{Bau2017}. Our results
indicate that removing the 2nd layer (in experiment L2) removes a higher-level
feature that is more predictive of the class label, thus resulting in a lower
accuracy.
In addition, by comparing the results from Figures \ref{fig:expr_sec_b2},
\ref{fig:expr_sec_c13} and \ref{fig:expr_sec_c24}, we can
observe that even an impoverished CVNN can produce significant gains over a
RVNN.


On the other hand, ablating the convolutional filters produces a different
behavior from ablating the outputs. Figures \ref{fig:c3_st3} (bottom) and
\ref{fig:c4_ost2} (bottom) show that the results are mixed. In some cases,
removing the second layer's imaginary filters produce the highest accuracy (e.g.
Figure \ref{fig:c3_st3}) while in other cases, removing the first layer's
imaginary filters produce the highest accuracy (e.g. Figure \ref{fig:c4_ost2}).
We speculate that the additional cross-terms in CVNNs produce redundant
information that can cause overfitting in some cases; our ablations removed this
redundancy which in turn improved accuracy. In many cases, removing the
convolutional layers produces a great range of differences in accuracy than
removing the output layers; this difference indicates that the convolutional
layers play a more important role in terms of predictive accuracy than the
outputs for CVNNs.

\section{Conclusion}
\label{sec:conc}
CVNNs are a more accurate model for wireless device classification than RVNNs, with CVNNs consistently outperforming RVNNs with approximately the same number of parameters. The main benefit that CVNNs provide over RVNNs is a more effective use of the joint information between in-phase and quadrature components of the signals, with this joint information being captured by cross-terms produced by complex number operations. Finally, we also showed through ablation experiments that deeper layers of CVNNs capture important higher-level features that can improve accuracy. 



\bibliographystyle{IEEEtran}
\bibliography{07.Ref}

\end{document}